\documentclass{article}

\usepackage{PRIMEarxiv}

\usepackage[utf8]{inputenc} 
\usepackage[T1]{fontenc}    
\usepackage{hyperref}       
\usepackage{url}            
\usepackage{booktabs}       
\usepackage{amsfonts}       
\usepackage{nicefrac}       
\usepackage{microtype}      
\usepackage{lipsum}
\usepackage{fancyhdr}       
\usepackage{graphicx}       
\graphicspath{{media/}}     

\usepackage{authblk}
\usepackage{booktabs}
\usepackage{xspace}
\newcommand*{\eg}{e.g.\@\xspace}
\newcommand*{\ie}{i.e.\@\xspace}
\newcommand*{\etal}{et al\@\xspace}
\usepackage{pifont}
\newcommand{\cmark}{\ding{51}}%
\newcommand{\xmark}{\ding{55}}%
\usepackage{color}
\usepackage{array}
\usepackage{amsmath}

\usepackage[linesnumbered,ruled,vlined]{algorithm2e}
\SetKwInput{KwInput}{Input}                
\SetKwInput{KwOutput}{Output}              
\SetKwInput{KwInitial}{Initialization}
\SetKwFor{For}{for}{}{}

\title{Visual Superordinate Abstraction for \\ Robust Concept Learning
}

\author[1]{Qi Zheng}
\author[2]{Chaoyue Wang}
\author[3]{Dadong Wang}
\author[2,1]{Dacheng Tao}
\affil[1]{University of Sydney, NSW 2008, Australia}
\affil[2]{JD Explore Academy}
\affil[3]{DATA61, CSIRO, NSW 2122, Australia}

\begin{document}
\maketitle


\begin{abstract}
Concept learning constructs visual representations that are connected to linguistic semantics, which is fundamental to vision-language tasks. Although promising progress has been made, existing concept learners are still vulnerable to attribute perturbations and out-of-distribution compositions during inference. We ascribe the bottleneck to a failure of exploring the intrinsic semantic hierarchy of visual concepts, e.g. \{red, blue,...\} $\in$ `color' subspace yet cube $\in$ `shape'. In this paper, we propose a visual superordinate abstraction framework for explicitly modeling semantic-aware visual subspaces (i.e. visual superordinates). With only natural visual question answering data, our model first acquires the semantic hierarchy from a linguistic view, and then explores mutually exclusive visual superordinates under the guidance of linguistic hierarchy. In addition, a quasi-center visual concept clustering and a superordinate shortcut learning schemes are proposed to enhance the discrimination and independence of concepts within each visual superordinate. Experiments demonstrate the superiority of the proposed framework under diverse settings, which increases the overall answering accuracy relatively by 7.5\% on reasoning with perturbations and 15.6\% on compositional generalization tests. 
\end{abstract}

\section{Introduction}
\label{sec:intro}

Concept learning actively extracts representations from a visual scene and connects them to linguistic tokens for identification, which resembles human cognition~\cite{inhelder2013early}. For example, we may focus on its color (\eg reddish-yellow) and shape (\eg round) appearance when identifying an orange. Concept learning is fundamental to vision-and-language tasks that perform multi-step inference over a scene's entities and their relationships, such as Visual Question Answering (VQA)~\cite{antol2015vqa}, Visual Commonsense Reasoning (VCR)~\cite{zellers2019recognition} and Vision-Language Navigation (VLN)~\cite{anderson2018vision}.

In recent years, great progress has been made on concept learning to increase the accuracy of identification. In the beginning, most algorithms require explicit annotations, such as ground truth scene graphs, for training concept learners~\cite{mascharka2018transparency,yi2018neural}.
 
\begin{figure}[th]
\centering
\includegraphics[width=.9\textwidth]{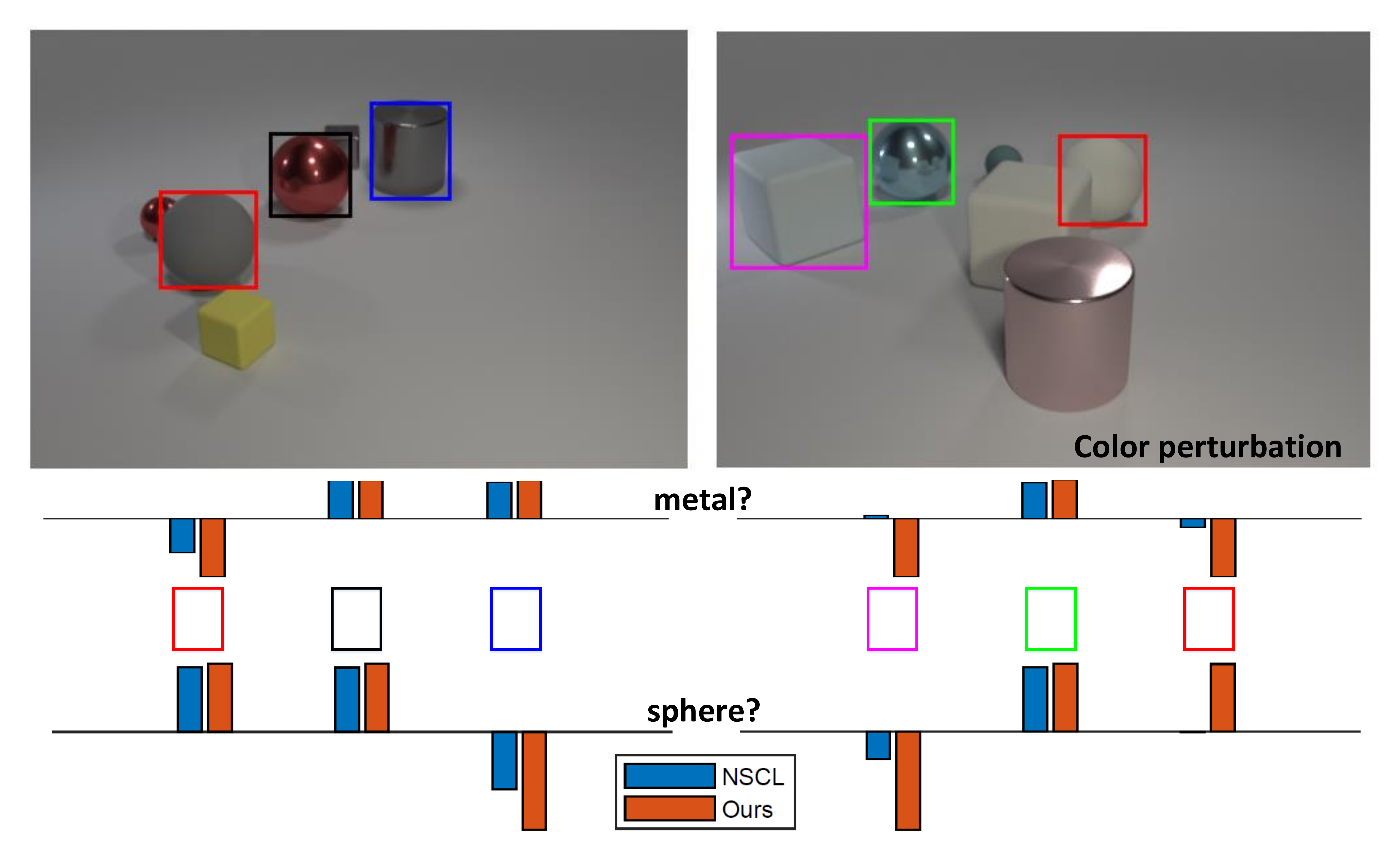}
\caption{Illustration of color perturbations. The left image share consistent attributes with training samples, and the right one is synthesized by adding perturbations to the color attribute. We present the scores of judging \textit{metal} and \textit{sphere} by NSCL~\cite{mao2019neuro} (in \textcolor{blue}{blue} bars) and our method (in \textcolor{red}{red} bars). The colorful rectangles refer to the object in corresponding bounding boxes.}
\label{fig:demon}
\end{figure} 

Until recently, Mao \etal~\cite{mao2019neuro} proposed quasi-symbolic execution together with learning visual subspaces using only question-answer pairs for images. Though the state-of-the-art methods can achieve around $99\%$ accuracy on diagnostic datasets such as the CLEVR dataset~\cite{johnson2017clevr}, they suffer from heavy performance drop on out-of-distribution compositions~\cite{mascharka2018transparency,yi2018neural,mao2019neuro,marois2018transfer}. Also, we observe that they are vulnerable to attribute perturbations during inference. 

A demonstration of attribute perturbation is shown in Figure~\ref{fig:demon}. The left scene shares exactly the same setting as training data, \ie object attributes such as color, shape, and material follow the same distributions with training data. In the right scene, small perturbations are added to the color attribute, \eg red$\to$light red. To examine a learner’s robustness to color perturbation, we output the scores of judging \textit{metal} and \textit{sphere} from the well-trained NSCL model~\cite{mao2019neuro} (shown in blue bars). We find it performs perfectly in the left scene, while much worse in the right one. It reveals that the learner’s recognition of the \textit{material} attribute and the \textit{shape} attribute are affected by \textit{colors}. We ascribe this phenomenon to a failure of exploring the intrinsic semantic hierarchy of visual concepts, \eg \{red, blue,...\} $\in$ `color' subspace yet cube $\in$ `shape'.

An abstraction of such a hierarchy is vital to humans' reasoning. It is one of the fundamental capabilities that make us being robust to understand the real world and generalize well to unseen scenarios~\cite{inhelder2013early,murphy2004big,landauer1997solution,lund1996producing,lake2020word,tenenbaum2011grow,rosch1976basic,tanaka1991object}. Human learners capture the essential hypothesis spaces in parsimonious form, and the formed hierarchy structure enables describing not only the specific situation at hand, but also a broader class of situations over which learning should generalize~\cite{tenenbaum2011grow}. As an attempt to introduce such hierarchy into machine learning, Han \etal~\cite{han2020visual} propose a visual concept-metaconcept learner (VCML) that utilizes extra metaconcept supervision. Different from them, we learn the semantic hierarchy with only natural VQA data. To the best of our knowledge, we are the first to explore the intrinsic semantic hierarchy under natural weak supervision.

In this paper, we propose a visual superordinate abstraction framework that explicitly models semantic-aware visual subspaces, which are denoted as visual superordinates. With the weak supervision from visual question answering data, the concept learner first acquires the semantic hierarchy from a linguistic view in a simple curriculum. Then in the following curriculum, it explores mutually exclusive visual superordinates under the guidance of linguistic hierarchy. Upon the abstraction framework, we propose a quasi-center visual concept clustering and a superordinate shortcut learning schemes, which further enhances the discrimination and independence of concepts within each visual superordinate. 
Quasi-center visual concept clustering aims to model the relationships between the clusters of visual representations and the linguistic features. By introducing a quasi-center of the visual cluster, we simultaneously (i) reduce the distance between visual representations and the quasi-center, and (ii) increase the similarity between visual samples and the corresponding concept. As for superordinate shortcut learning, the learner rectifies its judgment by reducing spurious causal-effect relations among superordinates.

Finally, comprehensive experiments are conducted to verify the effectiveness and generalization ability of the proposed concept learner. For the demonstration in Figure~\ref{fig:demon}, our method identifies metal and sphere correctly even with color perturbations. Statistically, the proposed model achieves comparable accuracy as the state-of-the-art methods. On the more challenging CLEVR-Perturb setting, our method outperforms NSCL~\cite{mao2019neuro} by a relative $7.5\%$ improvement. On the CLEVR-CoGenT dataset, our model overcomes the implicit bias in the training data, and performs the best without finetuning on split val-B, which surpasses the state of the art by $15.6\%$.

\section{Related Work} \label{sec:re_w}
\paragraph{Concept learning} Recent exploration on elementary visual reasoning starts from~\cite{johnson2017clevr}, which provides a diagnostic dataset to test the reasoning and generalization ability of a model by answering visual-related questions. As for joint learning of vision and natural language, existing methods mainly diverse in visual representations and question parsing process. Initial methods conduct reasoning primarily on convolutional feature maps. Johnson \etal~\cite{johnson2017inferring} build a reasoning system composed of a program generator and an execution engine, where the engine executes the decoded program sequence on features maps obtained from CNNs~\cite{he2016deep}. Hu \etal~\cite{hu2017learning} propose end-to-end module networks that conduct reasoning by directly predicting instance-specific network layouts without off-the-shelf parser. To further get rid of annotated layout data, Hu \etal~\cite{hu2018explainable} replace the layout graph in~\cite{hu2017learning} with a stack-based data structure that allows fully differentiable optimization. Similarly, Mascharka \etal~\cite{mascharka2018transparency} propose a set of visual reasoning primitives that composes a model according to a given question.  

Different from the methods reasoning on convolutional feature maps, Yi \etal~\cite{yi2018neural} parse a scene into a structural graph using trainable object and attribute detectors. Since the scene parser in~\cite{yi2018neural} requires ground-truth structural graphs for training, Mao \etal~\cite{mao2019neuro} propose quasi-symbolic execution and simultaneously learn scene parser and semantic parser using only question-answer pairs for images. Li \etal~\cite{li2020competence} propose a multi-dimensional Item Response Theory (mIRT) model for guiding the learning process with an adaptive curriculum to increase training efficiency. Prerez \etal~\cite{perez2018film} design a feature-wise linear modulation layer to improve the reasoning ability of the vanilla baseline in~\cite{johnson2017inferring}. Hudson and Manning~\cite{hudson2018compositional} introduce a recurrent memory, attention, and composition (MAC) cell that maintains control and memory separately to balance transparency and versatility, not explicitly parsing questions into programs. Following MAC~\cite{hudson2018compositional}, Wang \etal~\cite{wang2021interpretable} devise an object-centric compositional attention model to induce symbolic concept space. To boost the quality of detected objects, Kamath \etal~\cite{kamath2021mdetr} pre-train an end-to-end modulated detector that detects objects in an image conditioned on a raw text query, and then finetune on visual reasoning datasets. We follow the neuro-symbolic reasoning process in \cite{mao2019neuro}, but different from their semantic-agnostic visual representations, we learn visual superordinates to increase the robustness.

\paragraph{Causal inference} 

Causal analysis infers probabilities under both static conditions and changing conditions, which aims to figure out correlations in biased data~\cite{pearl2009causal}. Structural causal model (SCM), a directed graph that reveals causal relationships between random variables, has been developed as a formal tool to model causation from statistical data and counterfactual reasoning. It has been widely used in medical, psychological, and social research~\cite{dunn2015evaluation,king2008political,mackinnon2007mediation,richiardi2013mediation} to determine the effect of a treatment or policy, and recently been introduced into computer vision~\cite{nair2019causal,niu2020counterfactual,qi2020two,wang2020visual,yang2020deconfounded,tang2020unbiased} to enable counterfactual reasoning.

Though efficient, it can hardly be used in concept learning models without semantic-aware visual subspaces. In a flat concept set, the distance between any two concepts is not comparable with that of another two, neither the distance between visual features and different concepts. For example, an object representation can be close to both ``red’’ and ``cube’’, which gives no hint to its distance to other concepts. However, within a visual superordinate, the closeness to ``red’’ indicates a relatively large distance to ``green’’. Our visual superordinate abstraction framework models semantic-aware visual subspaces as independent variables, which enables the advanced strategies to be introduced in concept learning.

\begin{figure*}[t]
\centering
\includegraphics[width=.95\textwidth]{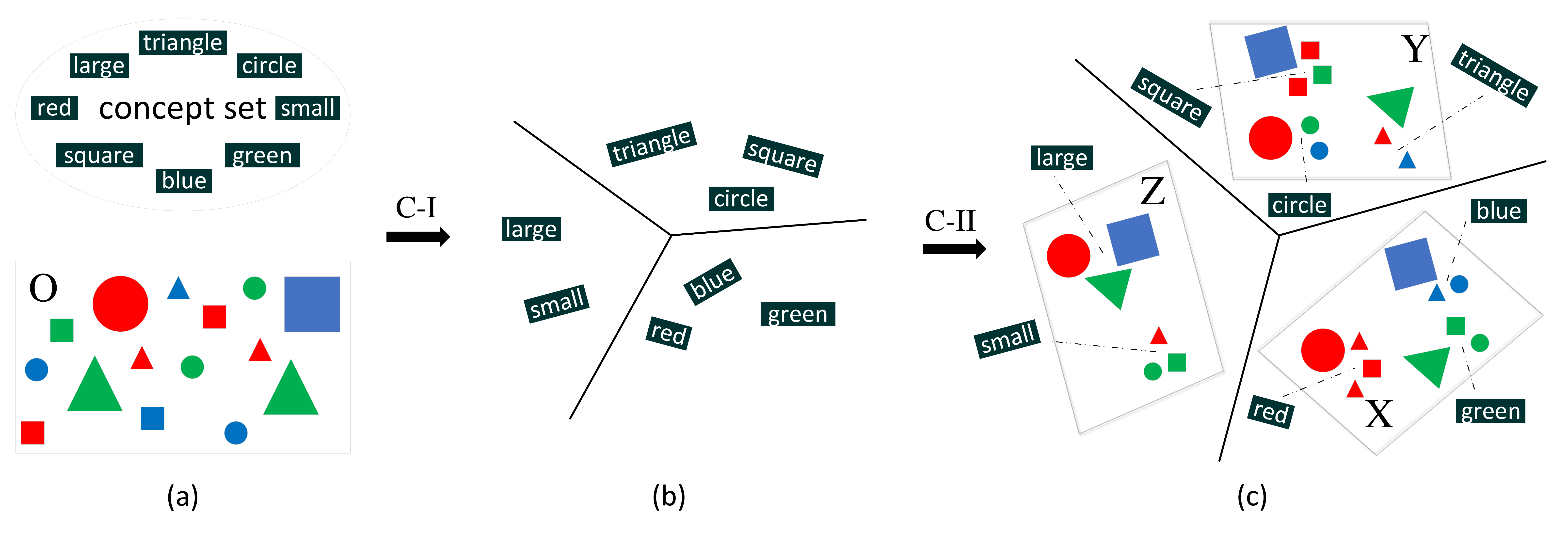}
\caption{{\bf An overview of the learning process of the proposed Visual Superordinate Abstraction framework. (a)} Originally, there is a set of visual concepts and detected objects from visual question answering data. {\bf (b)} After a simple curriculum (C-I), the learner acquires linguistic abstraction from the weak supervision. It categorizes concepts (\eg `red', `circle', `small') into different linguistic hierarchy (\eg \textit{color}, \textit{shape} and \textit{size}). {\bf (c)} In the following difficult curriculum (C-II), the learner constructs semantic-aware subspaces (\ie visual superordinates $X$, $Y$ and $Z$) aligning with the linguistic hierarchy. In each subspace, it extracts only one desired visual feature from detected objects, which can be described with the closest concept. For forming ideal visual superordinates, quasi-center visual concept clustering (Figure~\ref{fig:contrastive}) and superordinate shortcut learning (Figure~\ref{fig:causal}) are proposed and introduced in the following.}
\label{fig:framework}
\end{figure*}

\section{Method}
To conduct neuro-symbolic reasoning, the pipeline generally includes a concept learner and two parsers for vision and language respectively. Following NSCL~\cite{mao2019neuro}, in our method, visual parsing is done via pre-trained object detectors. The language parser converts a given question into a sequence of programs that consist of concepts and operations. For instance, the question ``What is the size of the sphere left of the blue metal object?'' is parsed into the program sequence Filter ($<$ObjConcept 2 (blue metal)$>$, Scene) $\to$ Filter ($<$RelConcept 1 (left)$>$,$\cdot$) $\to$ Filter($<$ObjConcept 1 (sphere)$>$,$\cdot$) $\to$ Query($<$Attribute 1 (size)$>$,$\cdot$). The key reasoning step is judging if an object shares a concept.

\subsection{Visual Superordinate Abstraction}

\paragraph{Linguistic hierarchy}
Let $\{\mathbf{v}_1, \mathbf{v}_2, \dots\}$ denotes the visual features of objects in a scene, $\{{c}_1, {c}_2, \dots, c_M\}$ is the concept set to be learned. These concepts describe $K$ attributes in total, yet it is unknown which concept belongs to which attribute. The concept learner learns $K^*$ mapping functions $f_k$, and forms $K^*$ visual subspaces $\{\mathcal{F}_k\}_{k=1}^{K^*}$, each $\mathcal{F}_k=\{f_k(\mathbf{v}_i)|\mathrm{for~all~} \mathbf{v_i}\}$. 

In previous methods, $\mathcal{F}_k$ are semantic-agnostic visual subspaces, thus the probability of the $i$-th object sharing $c_m$ concept is estimated as,
\begin{align} \label{eq:fullprob1}
    p(\mathbf{v}_i \mathrm{~has~} c_m)&{=}\sum_{k=1}^{K^*} p(c_m {\to} \mathcal{F}_k) p(\mathbf{v}_i \mathrm{~has~} c_m | c_m {\to} \mathcal{F}_k)  \nonumber \\
&{=}\sum_{k=1}^{K^*}b_m^{k}\sigma((f_k(\mathbf{v}_i)^T \mathrm{E}_k(c_m) - \gamma) / \tau),
\end{align}
where $\mathrm{E}_k(c_m)$ is the embedding of concept $c_m$ corresponding to $\mathcal{F}_k$ and $\sigma(\cdot)$ is the sigmoid function. $\gamma$ and $\tau$ are the shifting and the scaling parameters respectively. 
The prior probability $p(c_m{\to} \mathcal{F}_k)$ is given by the normalized vector $\mathbf{b}_m=[b_m^1,\dots,b_m^{K_1}]$ along with the concept $c_m$. The conditional probability $p(\mathbf{v}_i \mathrm{~has~} c_m | c_m {\to} \mathcal{F}_k)$ is given by the cosine similarity in the $k$-th subspace, where the normalization denominator is omitted.

The goal of our method is to first learn the linguistic hierarchy that naturally exists in visual reasoning questions. 
Different from the curriculum in NSCL~\cite{mao2019neuro} that arranges lessons by the depth of programs, we devise the curricula from easy question types to hard ones. For curriculum-I, we use programs of depth less than six and scenes including objects less than six, where the \textit{query}-type question-answer pairs contain soft alignment of linguistic hierarchy. Our experiments show that, after the curriculum-I, our model is capable of accurately defining linguistic hierarchy, \ie the affiliation illustrated in Figure~\ref{fig:framework} (b). With explored linguistic hierarchy, we are possible to further explore semantic-aware visual superordinates and their inside clustering. 


\noindent \textit{Training objectives.} Given parsed questions $q'$ from a pre-trained language parser, the concept learner is fully differentiable and trained by maximizing the likelihood of right answers
\begin{align}
\mathcal{L}_1(\theta)=\mathbb{E}_{(I,q',a)}[-\log p(\mathrm{exe}(I,q';\theta))=a],
\end{align}
where $\mathrm{exe}(I,q';\theta)$ represents the quasi-symbolic program execution and $\theta$ is the set of parameters in the concept learner. The execution result of each step is given by Eq.~(\ref{eq:fullprob1}).


\noindent \textbf{Semantic-aware visual superordinate abstraction.}\\
Though the linguistic hierarchy is learned, the visual mappings $f_k$ are still semantic-agnostic. It leads to entanglement between visual representations. The cause lies in the full-probability reasoning given by Eq.~(\ref{eq:fullprob1}). It indicates that the learner must refer to each subspace to determine whether an object shares the concept $c_m$. However, as discussed in our Introduction, we argue that $c_m$ should exclusively belong to only one visual subspace. 

To this end, we utilize the acquired linguistic hierarchy to learn semantic-aware visual subspaces, \ie the visual superordinates $X,~Y,~Z$ in Figure~\ref{fig:framework} (c). Upon the learned linguistic hierarchy, the concept learner can uniquely relate $c_m$ to visual subspace $X$ (or $Y$ or $Z$) given by $\arg\max_k b_m^k$, and estimate the probability of the $i$-th object sharing $c_m$ concept. For example, we assume that a concept $c_m$ belongs to subspace $X$, then
\begin{align} 
p(\mathbf{v}_i \mathrm{~has~} c_m) = \frac{e^{(f_X(\mathbf{v}_i)^T \mathrm{E}_X(c_m) - \gamma) / \tau}}{\sum_{c_{m'}\to F_X}e^{(f_X(\mathbf{v}_i)^T \mathrm{E}_X(c_{m'}) - \gamma) / \tau}}. \label{eq:posterior1}
\end{align}
Comparing Eq.~(\ref{eq:fullprob1}) and Eq.~(\ref{eq:posterior1}), we can see that $c_m$ uniquely matches $F_X$ and only refers to $F_X$ in judgment. 
This way, the subspaces $X$ become aware of the superordinate such as \textit{color} and \textit{shape}.

A similar objective is utilized to optimize the learner during the visual superordinate abstraction,
\begin{align}
\mathcal{L}_2(\theta)=\mathbb{E}_{(I,q',a)}[-\log p(\mathrm{exe}^*(I,q';\theta))=a],
\end{align}
where $\mathrm{exe}^*$ executes each program step and produces results by Eq.~(\ref{eq:posterior1}). 

Learning the visual superordinates increases the independence of different visual subspaces and improves the learner's robustness to perturbations. In addition, we devise a quasi-center visual concept clustering and a superordinate shortcut learning schemes to enhance the discrimination and independence of concepts within each visual superordinate.

\subsection{Quasi-center Visual Concept Clustering}
Given the independent mappings learned by visual superordinate abstraction, the visual representations within each superordinate can still be mixed with each other. For instance, the learner can distinguish \textit{shape} from \textit{color} and \textit{size}, but it may confuse \textit{cube}, \textit{cylinder} and \textit{sphere}. The reason can be found in Eq.~(\ref{eq:fullprob1}) and Eq.~(\ref{eq:posterior1}). The learner estimates the similarity between the mapped feature and the corresponding concept representation, yet ignores the samples that share the same concept in the visual subspace. As a result, visual samples diffuse around a concept and presents high variance with the concept representation, which is vulnerable to noise.

\begin{figure}[!t]
    \centering
    \includegraphics[width=.8\textwidth]{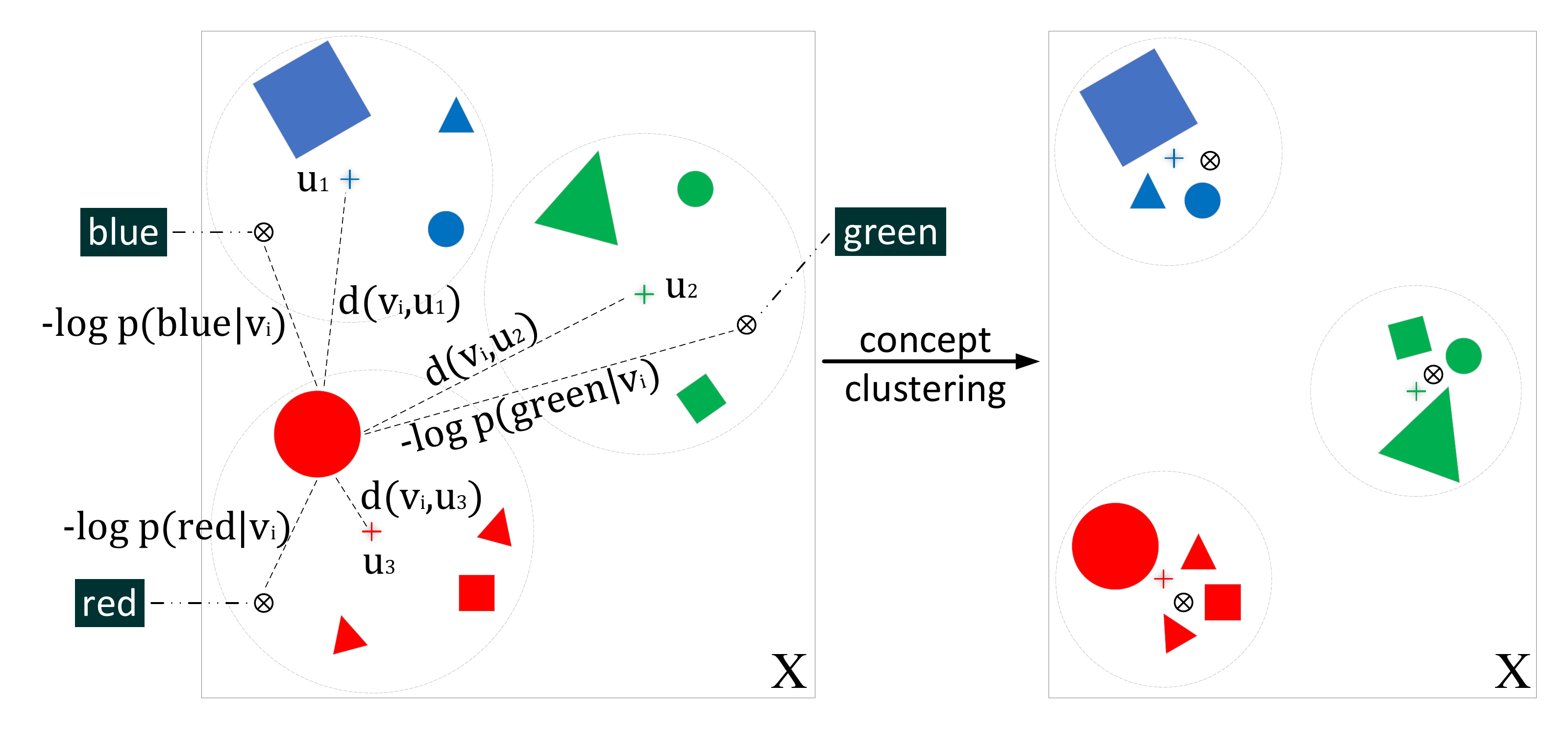}
    \caption{Demonstration of quasi-center visual concept clustering in \textit{color} superordinate. $\otimes$ denotes the projection of concept words in the visual subspace, and $+$ denotes the quasi-center of visual concept clusters (\ie $\mathbf{u}_m$). Quasi-center concept clustering not only accumulates the visual representations around a concept, but also reduces the distance between concept words (\eg $\otimes$ for ``red'') and corresponding cluster centers (\eg \textcolor{red}{$+$}). View the color version.}
    \label{fig:contrastive}
\end{figure}

As illustrated in Figure~\ref{fig:contrastive} (left), the visual representations around the color concepts may disperse within each cluster and the clustering center may also be far from the corresponding concept representation. To this end, we design a memory cache to enhance the clustering of visual features around different concepts and reduce the discrepancy between linguistic concepts and visual representations. Supposing that at the $t$-th training step the learner has seen a set of samples that share a concept $c_m$, a quasi-center of each concept cluster can be calculated as
\begin{align}
\mathbf{u}_m^{(t)}&=\frac{1}{|c_m^{(t)}|}\sum_{\mathbf{v}_i\in\{c_m^{(t)}\}}f_{k}(\mathbf{v}_i) \\
&=\frac{|c_m^{(t-1)}|\mathbf{u}_m^{(t-1)} + \sum_{\mathbf{v}_i\in{\{\Delta c_m^{(t)}\}}}f_{k}(\mathbf{v}_i)}{|c_m^{(t-1)}|+|\Delta c_m^{(t)}|}, \label{eq:incremental}
\end{align}
where $\mathbf{v}_i$ are the samples that shares $c_m$ up to $t$-th training step and $k$ is the superordinate that $c_m$ aligns with according the learned linguistic hierarchy. Eq.~(\ref{eq:incremental}) dynamically updates the clustering center at each training step, where $\Delta c_m=\{c_m\}^{(t)}\setminus\{c_m\}^{(t-1)}$ are the samples that appear at the $t$-th training step.

For simplicity, we calculate the Euclidean distance between a mapped sample and the existing clustering quasi-center
\begin{align}
d(\mathbf{v}_i, \mathbf{u}_m^{(t)})=\Vert f_{k}(\mathbf{v}_i)-\mathbf{u}_m^{(t)}\Vert_2,
\end{align}
and balance it with the original cosine similarity by
\begin{align} \label{eq:contra}
\log p^{'}(\mathbf{v}_i \mathrm{~has~} c_m){=}\log p(\mathbf{v}_i \mathrm{~has~} c_m) {-} \alpha d(\mathbf{v}_i, \mathbf{u}_m^{(t)}),
\end{align}
where $\alpha$ is the decay coefficient of the distance.

Adding cached visual samples, the learner infers the probability for concepts using Eq.~(\ref{eq:contra}) instead of Eq.~(\ref{eq:posterior1}). On the one hand, quasi-center visual concept clustering not only concentrates the distribution of visual representations. It reduces the distance between visual samples and the clustering center for a more consistent visual representation. On the other hand, the proposed clustering also updates the representation of concept words, which increases the similarity between (the clustering center of) visual samples and corresponding concepts. This produces a better multi-modal embedding and increases the robustness of concept identification, as demonstrated in Figure~\ref{fig:contrastive} (right).

\subsection{Superordinate Shortcut Learning}
\begin{figure}[!h]
    \centering
    \includegraphics[width=.8\textwidth]{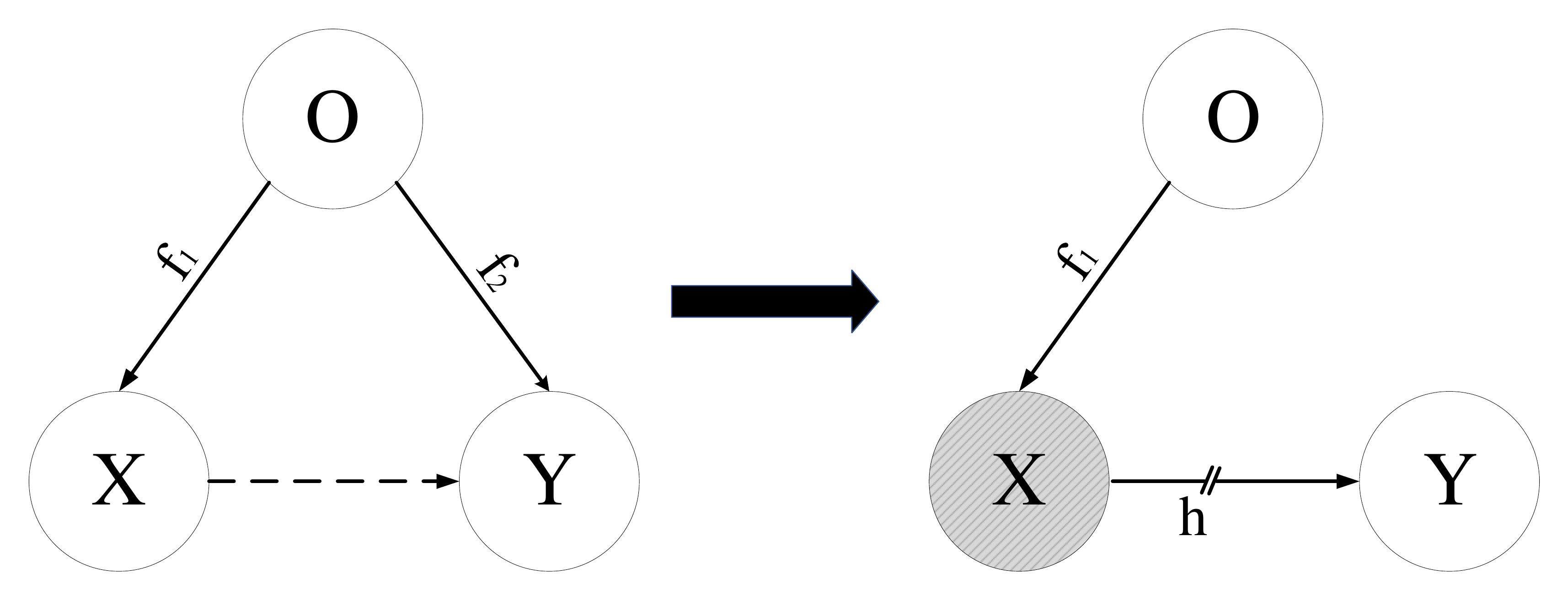}
    \caption{Demonstration of superordinate shortcut learning. To measure whether there is a spurious causal effect between $X$ and $Y$, the learner explicitly learns a shortcut $h$ by interfering $Y$ from $X$. ${//}$ is the stop-gradient operator, which indicates that no gradient propagates through $X$ when training $h$.}
    \label{fig:causal}
\end{figure}

Quasi-center concept clustering concentrates different visual clusters within a superordinate, meanwhile, we can analyze the dependence among superordinates. As show in Figure~\ref{fig:causal}, $X\doteq \{f_1(\mathbf{v}_i)\}$ and $Y\doteq \{f_2(\mathbf{v}_i)\}$ are the visual representations extracted from object $O$ for two different superordinates. Due to the expressiveness of modern neural networks, a learner is likely to build a causal effect between these two variables that are expected to be independent (dashed line in Figure~\ref{fig:causal}). Supposing that there is a correlation between $X$ and $Y$ in training data and the learner figures out a shortcut from $X$ to $Y$. Thus $Y$ is simultaneously influenced by $O$ and $X$. Given the learned visual superordinate abstraction, the learner can explicitly model the shortcut from $X$ to $Y$ by estimating $p(Y|X)$ during training. Then during inference, the true causal effect from $O$ to $Y$ can be recovered by blocking such a shortcut.

For an object $\mathbf{v}_i$ in $O$ space, the learner infers the probability of $\mathbf{v}_i$ sharing a concept corresponding to $X$ by Eq.~(\ref{eq:posterior1}), which gives the probability distribution $p(\mathbf{v}_i \mathrm{~has~} c_x)$ for all $c_x {\to} X$. Then the attribute of $\mathbf{v}_i$ in $X$ is given by $c_{x^*}$, $x^*=\arg\max_x p(\mathbf{v}_i \mathrm{~has~} c_x)$. Assuming that this is known to the concept learner (\ie the shaded $X$ in Figure~\ref{fig:causal}), it estimates the information from $X$ to $Y$ by
\begin{align} \label{eq:short}
    p(\mathbf{v}_i \mathrm{~has~} c_y|X){=}\frac{e^{(h_\varphi(\mathrm{E}_1(c_{x^*}))^T \mathrm{E}_2(c_y) - \gamma) / \tau}}{\sum_{c_{y'}{\to} Y}e^{(h_\varphi(\mathrm{E}_1(c_{x^*}))^T \mathrm{E}_2(c_{y'}) {-} \gamma) / \tau}}, 
\end{align}
where $\mathrm{E}_1(c_{x^*})$ is the embedding of $c_{x^*}$ corresponding to $X$ subspace, and $h_\varphi(\cdot)$ is the function that maps concepts for superordinate $X$ to superordinate $Y$.

When learning the shortcut from $X$ to $Y$, the learner answers questions using Eq.~(\ref{eq:short}) instead of Eq.~(\ref{eq:posterior1}). The resulted loss only updates $\varphi$, with $\theta$ fixed. After training, the learner conducts regular inference with Eq.~(\ref{eq:posterior1}) and infers the true causal effect from $O$ to $Y$ by subtracting $p(Y|X)$ from initial estimates given by Eq.~(\ref{eq:posterior1}).Algorithm 1 elaborates the training process of the proposed visual superordinate abstraction for robust concept learning. 

\begin{algorithm}[!ht]
\SetAlgoLined
\DontPrintSemicolon
\KwInitial{linguistic prior $\mathbf{b}$, visual mappings $f$, concept embeddings $\mathrm{E}$, curriculum $\{(I,q',a)_n\}^{l_{1,2}}$, visual mappings $h_{X\to Y}$ if learning superordinate shortcut} 
\BlankLine
\For{$(I,q',a)$ in curriculum lesson 1 \tcp*[f]{learn linguistic abstraction}}{
  \For{program $e$ in $q'$ \tcp*[f]{execute parsed question $q'$}}{
  calculate $p(I,e;\mathbf{b},f,\mathrm{E})$ using Eq.~(1); \tcp*[f]{run on scene objects}
  }
  predict answer $a'=\mathrm{accumulate}(p)$;\\
  update $\mathbf{b},f,\mathrm{E}$ with loss $\mathcal{L}_1=\mathbb{E}_{(I,q',a)}[-\log p(a'=a)]$;
}
\BlankLine
\For{$(I,q',a)$ in curriculum lesson 2 \tcp*[f]{learn visual abstraction}}{
  \For{program $e$ in $q'$}{
    \lIf{with concept clustering}{
      calculate $p(I,e;\mathbf{b},f,\mathrm{E})$ using Eq.~(8);
    }
    \lElseIf{train shortcut branch $X\to Y$}{
      calculate $p(I,e;\mathbf{b},h,\mathrm{E})$ using Eq.~(9);
    }
    \lElse{
      calculate $p(I,e;\mathbf{b},f,\mathrm{E})$ using Eq.~(3);
    }
  }
  predict answer $a'=\mathrm{accumulate}(p)$;\\
  \lIf{type($q'$)==count}{
    update $\mathbf{b},f, (h),\mathrm{E}$ with loss $\mathcal{L}_2=\mathbb{E}_{(I,q',a)}[(a'-a)^2]$;
  }
  \lElse{
    update $\mathbf{b},f, (h),\mathrm{E}$ with loss $\mathcal{L}_2{=}\mathbb{E}_{(I,q',a)}[{-}\log p(a'{=}a)]$;
    \tcp*[f]{e.g., type($q'$)==query}
  }
}
 \caption{Weakly-supervised learning for visual superordinate abstraction}
\end{algorithm}

\section{Experiments}
We first provide the details of implementation in Section~\ref{sec:implement}. Then we conduct experiments on the CLEVR dataset\footnote{\url{https://cs.stanford.edu/people/jcjohns/clevr}}~\cite{johnson2017clevr} to provide an overall comparison of the reasoning ability under regular setting with the state of the arts in Section~\ref{sec:reason}. We evaluate the generalization ability to new compositions on CLEVR-CoGenT dataset$^\textrm{\textcolor{red}{1}}$~\cite{johnson2017clevr} in Section~\ref{sec:biased} and to perturbations on CLEVR-Perturb test data in Section~\ref{sec:perturb}, followed by clustering visualization and sensitivity analysis.

\subsection{Implementation Details} \label{sec:implement}
Our baseline model strictly follows the design of NSCL~\cite{mao2019neuro} for a fair comparison. We only use images and question-answer pairs for training and adopt curriculum learning to train the learner. Since the semantic parser without program annotations in NSCL~\cite{mao2019neuro} has achieved nearly perfect parsing accuracy, we use the pre-trained parser and fix it during training our visual superordinate abstraction model.

We set the dimension of word embedding and positional embedding as $300$ and $50$ respectively. The dimension of visual attribute subspaces is $64$. ResNet34 is used to extract the general feature of objects in $256$-dim, and subspace mappings consist of one-layer linear projection. We adopt the AdamW optimizer~\cite{loshchilov2017decoupled} for training, with initial learning rate of $0.001$. The batch size for training is set as $32$. The scaling parameter and the shifting parameter are set to $0.25$ and $0.85$, respectively. As for concept clustering, we only add the most likely positive sample into the cache at each training step, and the minimum samples to start the clustering is set to $50$. The distance decay coefficient $\alpha$ is set to $0.01$, and sensitivity analysis is given in Section~\ref{sec:perturb}. The module for superordinate shortcut learning consists of two linear layers and a non-linear activation layer. We alternatively update $\varphi$ and $\theta$ at the same frequency during training.

\subsection{General Visual Reasoning} \label{sec:reason}
The CLEVR~\cite{johnson2017clevr} dataset contains $100$k rendered images and about one million automatically generated question-answer pairs. The question types include querying attributes, comparing attributes (or numbers), counting, and logic reasoning like existence. The combinations during training are the same as those in testing. We compare our model with both implicit and explicit reasoning methods.

\setlength{\tabcolsep}{3pt}
\begin{table*}[ht]
\centering
\begin{tabular}{ccccccc}
\toprule
 & FiLM~\cite{perez2018film} & TbD~\cite{mascharka2018transparency} & NSCL~\cite{mao2019neuro} & MDETR~\cite{kamath2021mdetr} & MAC~\cite{hudson2018compositional} & Ours \\ \hline
CLEVR & 97.6 & 99.1 & 98.9 & \textbf{99.7} & 98.9 & 99.1 \\
CoGenT val-A & 98.3 & 98.8 & 97.9 & \textbf{99.8} & 96.9 & 98.0 \\
CoGenT val-B & 78.8 & 75.4 & 74.1 & 76.7 & 79.5 & \textbf{91.9} \\
\bottomrule
\end{tabular}
\caption{Overall question answering accuracy (\%) on CLEVR and CLEVR-CoGenT datasets.}
\label{tab:clevr}
\end{table*}

The overall results are in Table~\ref{tab:clevr} (first row), and a more detailed comparison can be found in our supplementary material. The end-to-end pretraining method MDETR~\cite{kamath2021mdetr} achieves the highest accuracy, which indicates the superiority of a jointly trained object detector. Among the rest methods, TbD~\cite{mascharka2018transparency} obtains the same highest score as our model. Though flexible, the training of TbD~\cite{mascharka2018transparency} requires annotated programs for question parsing. For the approaches FiLM~\cite{perez2018film}, MAC~\cite{hudson2018compositional} and NSCL~\cite{mao2019neuro} that do not use extra annotations, our method performs slightly better than them, which verifies the efficacy of the visual superordinate abstraction framework in regular reasoning.

\subsection{Robustness to New Compositions} \label{sec:biased}
\begin{figure}[h]
\centering
\includegraphics[width=\textwidth]{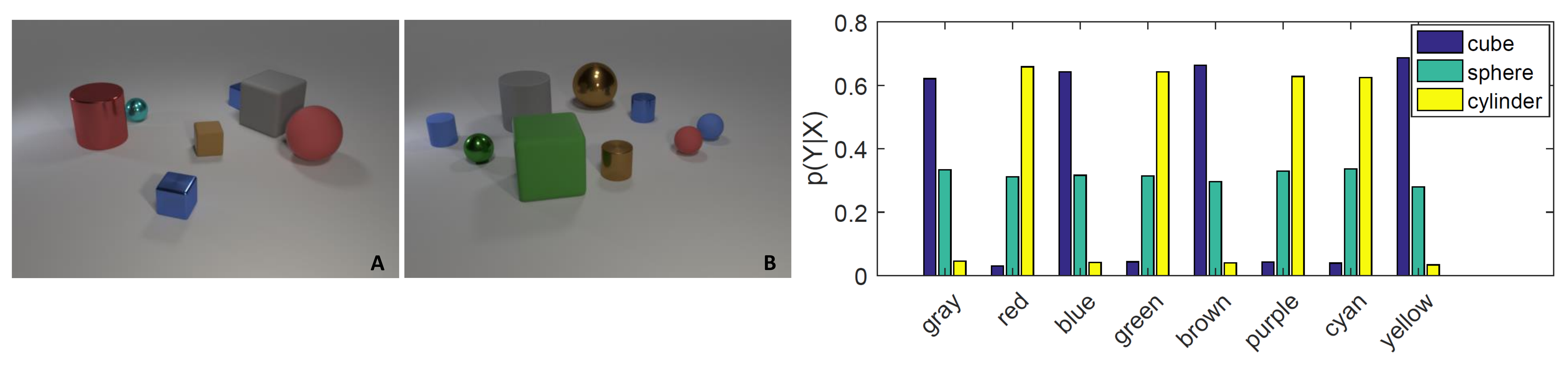}
\vspace{-0.2cm}
\caption{Illustration of biased data. Image A shows an example from CLEVR CoGenT val-A split, and image B is from the val-B split. The bar graph shows the learned correlation of $p(Y|X)$ from the biased training data.}
\label{fig:bias}
\end{figure}

The bias in training data can heavily affect the independence of learned concepts. CLEVR-CoGenT~\cite{johnson2017clevr} is proposed for a diagnose. There are two conditions for data splits: in Condition A all cubes are gray, blue, brown, or yellow and all cylinders are red, green, purple, or cyan; in Condition B these shapes swap color palettes. Both conditions contain spheres of all eight colors. Training split is under condition A and validation splits A and B are under two conditions respectively, as shown in Figure~\ref{fig:bias}.

In Figure~\ref{fig:causal}, by interfering $X$, the learner makes the most of the information transmitted in the shortcut from $X$ to $Y$. Interestingly, the learned shortcut plotted in Figure~\ref{fig:bias} is consistent with the bias in the training data. Specifically, given the color in \{gray, blue, brown, yellow\}, the learner predicts a high possibility (about $2/3$) of the cube shape, approximately twice that of the sphere (about $1/3$). The cylinder has a very small showing chance (close to zero). The opposite effect \textit{w.r.t} cube and cylinder can be observed in the rest four colors. If we assume that the shape is uniformly distributed, \ie $N_{cube}{=}N_{sphere}{=}N_{cylinder}$ in the training data, then the biased distribution of $p(\mathrm{shape}|\mathrm{color})$ is exactly $(1/3,~2/3)$ (or vise versa) for the colors appearing in the two splits. The learner is ready to use the disclosed bias in the inference on split val-B.

Overall statistic comparison with the state-of-the-art methods is listed in Table~\ref{tab:clevr}. Due to the distribution discrepancy, almost all the methods present an excellent reasoning ability on the val-A split, followed by a plunge on val-B. The val-A split data is consistent with CoGenT training data, where the color attribute correlates with the shape attribute. In this case, MDETR~\cite{kamath2021mdetr} and TbD~\cite{mascharka2018transparency} perform slightly better than the rest methods, including ours. On the val-B split, the correlation changes inversely, and all the comparing approaches are negatively affected. On the contrary, our model still obtains high accuracy without any fine-tuning on the val-B split. It attributes to learning a superordinate shortcut to reduce the spurious causal effect in training data.

\setlength{\tabcolsep}{1.5pt}
\begin{table*}[ht]
\centering
\begin{tabular}{l|c|c|c||c|c|c|c|c|c|c|c}
\toprule
& abs & cc & sl & Overall & Count & Exist & Cnt ($=$) & Cnt ($>$) & Cnt ($<$) & Comp. Attr. & Query \\ \hline
NSCL & \xmark & \xmark & \xmark & 74.1 & 71.8 & 85.7 & 72.5 & 82.2 & 80.8 & 80.7 & 66.7\\
Ours & \cmark & \xmark & \xmark & 77.7 & 75.0 & 87.6 & 74.5 & 84.2 & 82.5 & 87.6 & 70.1\\
Ours & \cmark & \cmark & \xmark & 78.5 & 76.6 & 88.0 & 77.5 & 84.2 & 82.5 & 87.7 & 70.8\\
Ours & \cmark & \cmark & \cmark & \textbf{91.9} & \textbf{95.7} & \textbf{99.1} & \textbf{94.0} & \textbf{98.8} & \textbf{98.7} & \textbf{99.1} & \textbf{81.5}\\
\bottomrule
\end{tabular}
\vspace{0.1cm}
\caption{Complete comparison on the val-B split of CLEVR-CoGenT dataset. Questions of comparing numbers are divided into count-equal (\ie Cnt(=)), count-greater-than (\ie Cnt($>$)), and count-less-than (\ie Cnt($<$)). ``abs'', ``cc'' and ``sl'' are abbreviations for \textit{visual superordinate abstraction}, \textit{quasi-center concept clustering} and \textit{superordinate shortcut learning} respectively.
}
\label{tab:cogent}
\end{table*}

Detailed comparison with NSCL~\cite{mao2019neuro} on the val-B split of CLEVR-CoGenT is provided in Table~\ref{tab:cogent}. Under biased training, NSCL perform worse on new compositions of color and shape. However, due to shortcut learning upon the visual superordinate abstraction, our model achieves higher accuracy on all types of questions, with a relative $4.9\%$ overall gain. Among the tasks, our model surpasses NSCL by a large margin (about $8.6\%$) on comparing attributes. Causal inference brings further relative $18.3\%$ overall improvement for our model. Significantly, it promotes the performance on counting by $27.6\%$, and reasons nearly perfectly on existence and comparing attributes.

\begin{figure}[h]
    \centering
    \includegraphics[width=.8\textwidth]{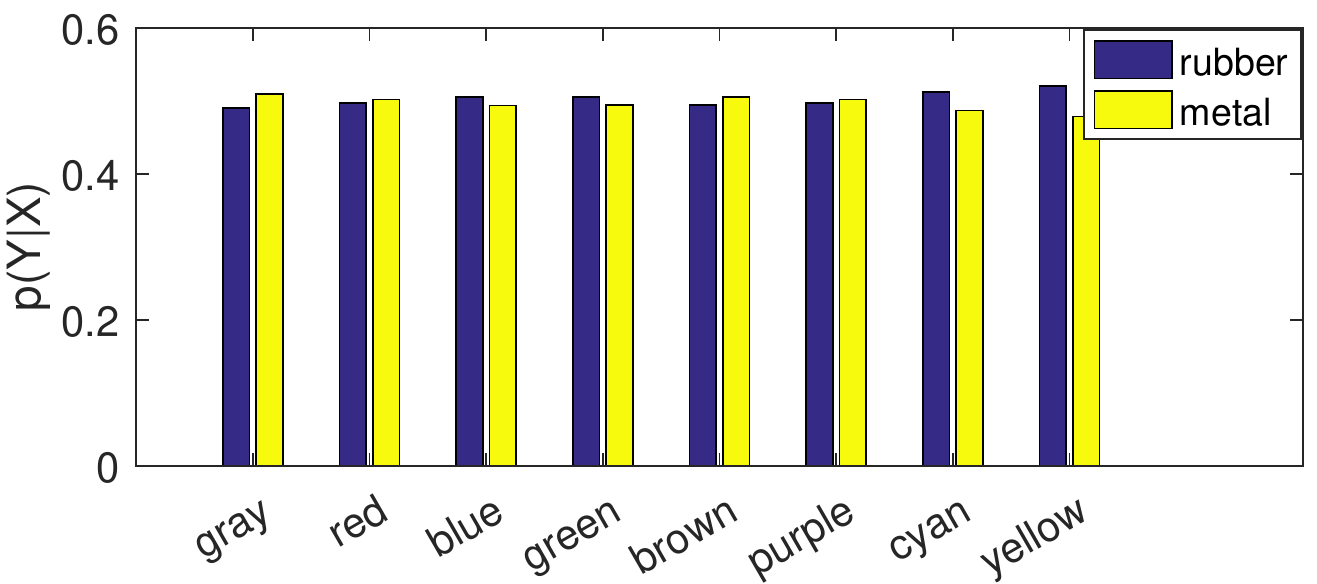}
    \caption{The learned correlation $p(Y|X)$ between \textit{color} and \textit{material} superordinates from the training data.}
    \label{fig:xto1}
\end{figure}

In Figure~\ref{fig:bias}, we analyze the correlation between \textit{color} and \textit{shape} in CLEVR-CoGenT training data. Here we train a shortcut from \textit{color} to \textit{material} in a similar way. The learned conditional probability distribution for different colors is provided in Figure~\ref{fig:xto1}. The result reveals that there is no bias related to this pair of variables, which is consistent with the ground-truth setting in CLEVR-CoGenT training set.

\subsection{Robustness to Perturbations} \label{sec:perturb}
\begin{figure}[h]
\centering
\includegraphics[width=.8\textwidth]{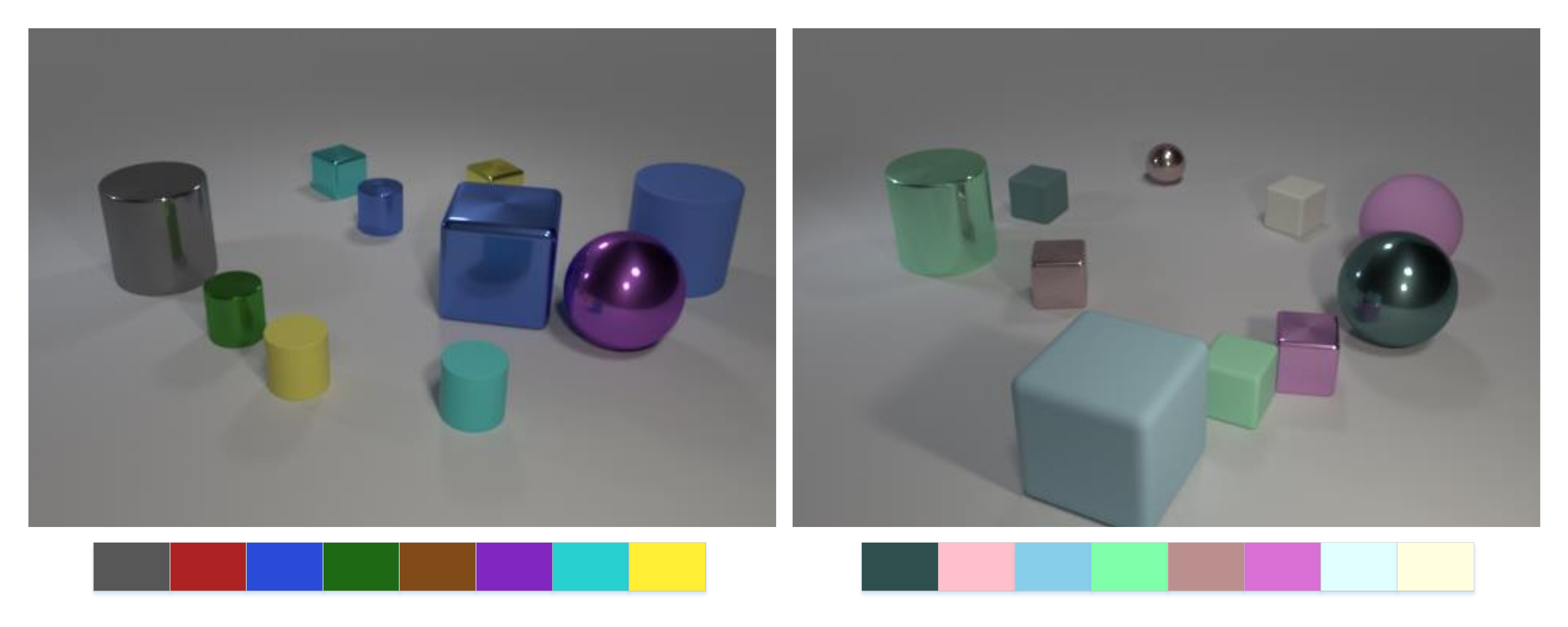}
\caption{Illustration of color perturbation. The left shows original image from CLEVR~\cite{johnson2017clevr}, and the right one is synthesized by adding color perturbations. The specific colors used in two settings are below the two images respectively.}
\label{fig:color}
\end{figure}

To examine the learner's robustness to perturbations in one superordinate, we synthesize a CLEVR-Perturb test set that contains $4$k images, each with about $10$ question-answer pairs. The values for shape, size and material are set the same as those in the CLEVR dataset~\cite{johnson2017clevr}. As for the colors, each color is slightly perturbed to a nearby value. The specific color shift is listed in Figure~\ref{fig:color}. To precisely assess the robustness to color perturbations, we only generate the questions that require no reasoning ability about color. For example, in the original setting, a possible question is ``How many other things are there of the same shape as the tiny cyan matte object?''. Though the final step is querying \textit{shape}, this question requires identifying ``cyan'' and will not appear in the synthesized CLEVR-Perturb test set. We compare our model with NSCL~\cite{mao2019neuro} that performs best on the CLEVR dataset without extra annotations, and with the baseline without concept clustering.

\setlength{\tabcolsep}{1.5pt}
\begin{table*}[ht]
\centering
\begin{tabular}{l|c|c|c||c|c|c|c|c|c|c|c}
\toprule
& abs & sl & cc & Overall & Count & Exist & Cnt ($=$) & Cnt ($>$) & Cnt ($<$) & Comp. Attr. & Query \\ \hline
NSCL & \xmark & \xmark & \xmark & 87.1 & 76.2 & 91.8 & 81.1 & 91.1 & 92.2 & 90.9 & 91.7 \\
Ours & \cmark & \xmark & \xmark & 92.9 & 85.5 & 96.6 & 87.8 & 96.0 & 95.8 & \textbf{95.7} & 95.7 \\
Ours & \cmark & \cmark & \xmark & 92.7 & 85.7 & 96.1 & 88.3 & 96.1 & \textbf{96.0} & 95.2 & 95.5 \\
Ours & \cmark & \cmark & \cmark & \textbf{93.6} & \textbf{87.3} & \textbf{96.9} & \textbf{89.0} & \textbf{96.5} & \textbf{96.0} & \textbf{95.7} & \textbf{96.0} \\
\bottomrule
\end{tabular}
\caption{Comparison on the CLEVR-Perturb test set. Questions of comparing numbers are divided into count-equal (\ie Cnt(=)), count-greater-than (\ie Cnt($>$)), and count-less-than (\ie Cnt($<$)). ``abs'', ``cc'' and ``sl'' are abbreviations for \textit{visual superordinate abstraction}, \textit{quasi-center concept clustering} and \textit{superordinate shortcut learning} respectively. }
\label{tab:perturb}
\end{table*}

Table~\ref{tab:perturb} lists the statistical results. Comparing the variant model without concept clustering and NSCL~\cite{mao2019neuro}, we can see obvious relative improvements on all types of questions (about $12.2\%$ for counting, $5.7\%$ for comparing numbers, $5.2\%$ for existence, $5.3\%$ for comparing attribute, and $4.4\%$ for querying). It indicates that applying linguistic abstraction leads to enhanced visual abstraction. Equipped with concept clustering, our model further boosts the performance on counting related questions, which shows the superiority of the proposed visual superordinate abstraction framework.

Referring to the qualitative example shown in Figure~\ref{fig:demon}, the well-trained NSCL model performs nearly perfectly on the left scene, while much worse on the right scene. It reveals that the learner’s recognition of the \textit{material} attribute and the \textit{shape} attribute are affected by \textit{colors}. Instead, our model identifies metal and sphere correctly. More importantly, we can observe that our model produces a nearly equivalent possibility for different objects for each concept. It indicates that the learner abstracts the feature that is only relevant to the current concept (\eg metal), so the objects that have the same material get a similar score. More examples can be found in the supplementary material.

\begin{figure*}[]
    \centering
    \begin{tabular}{ccc}
        \includegraphics[width=.32\textwidth]{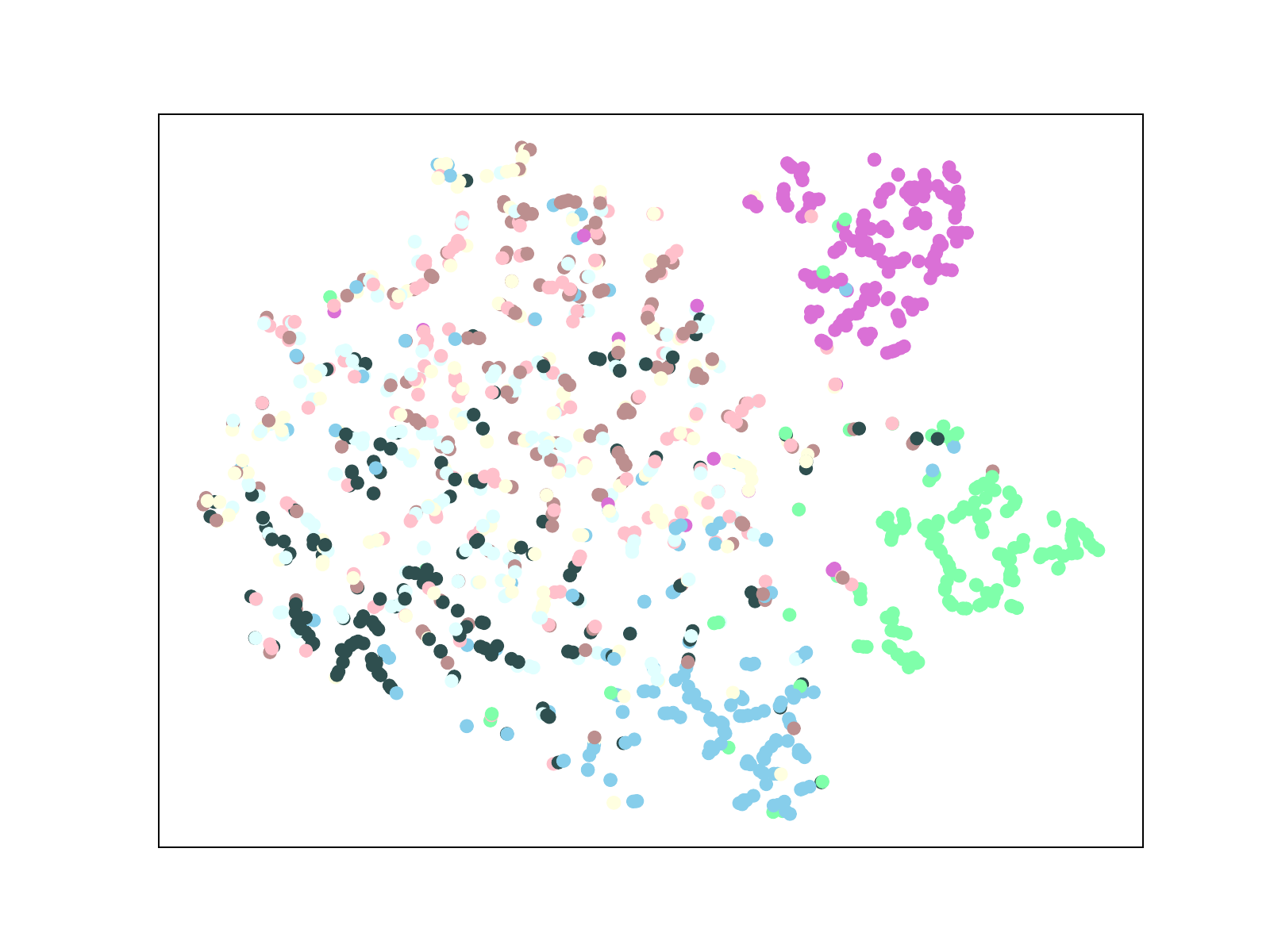} & \includegraphics[width=.32\textwidth]{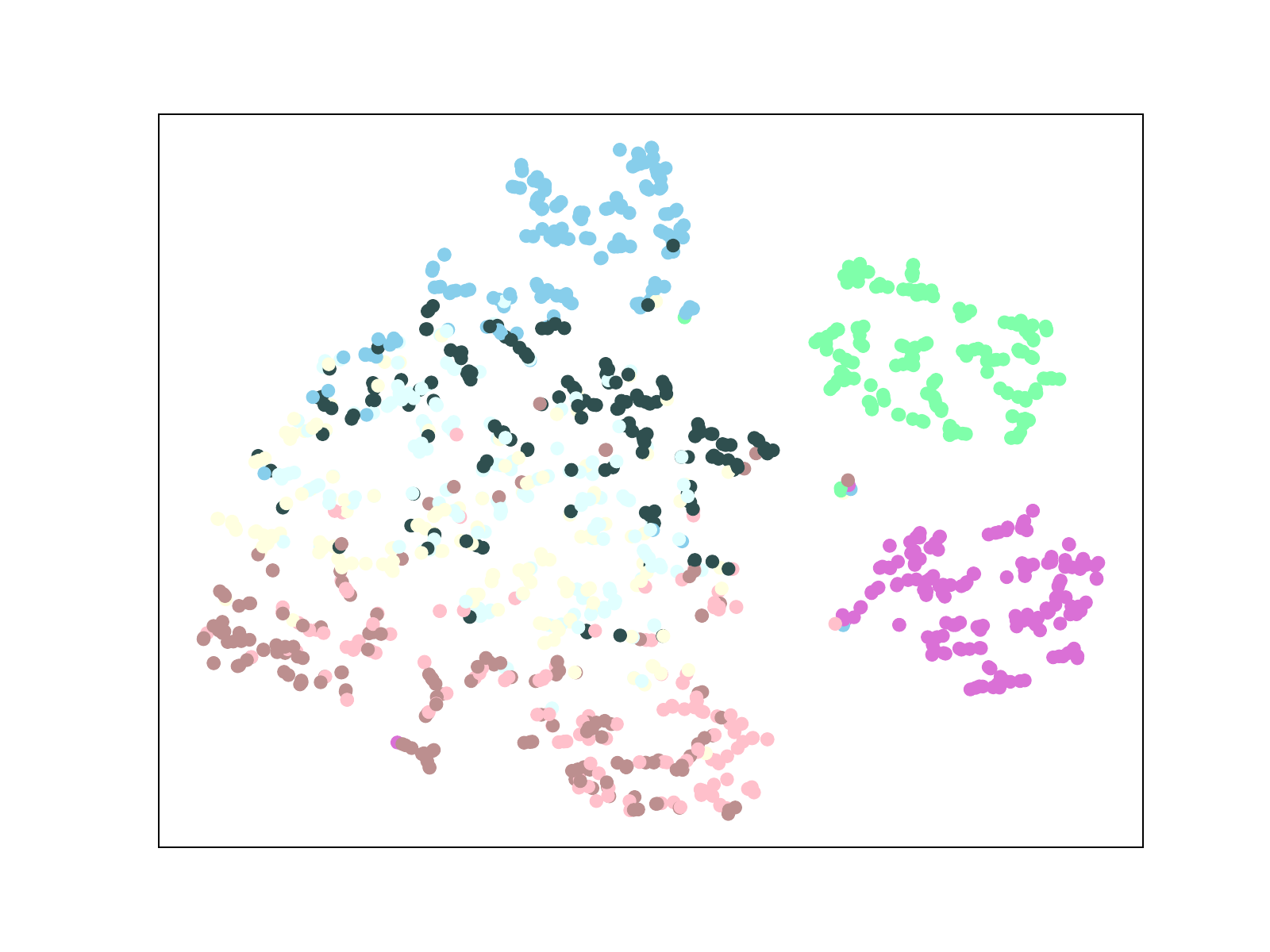} & \includegraphics[width=.32\textwidth]{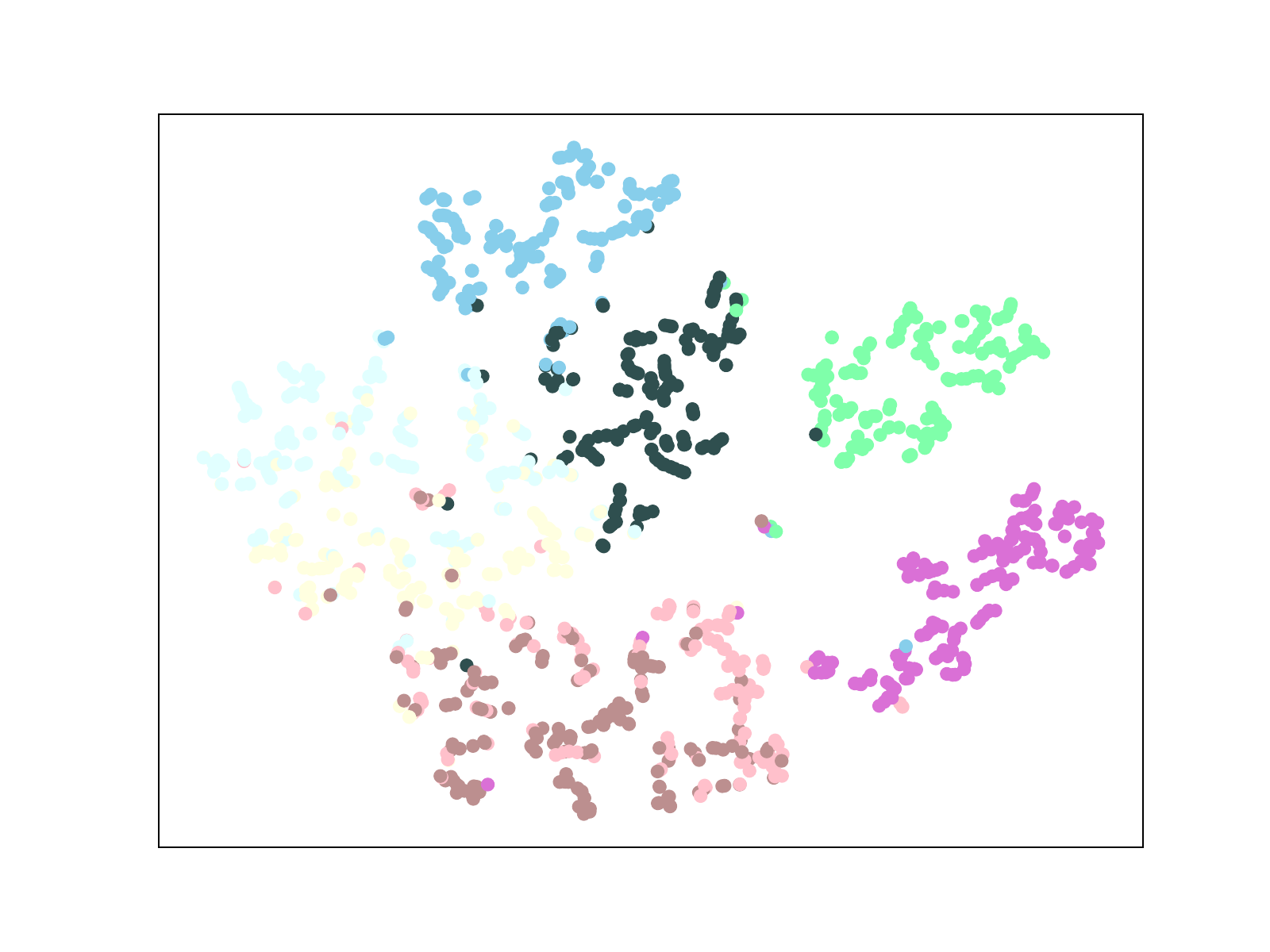} \\
        (a) NSCL~\cite{mao2019neuro} & (b) Ours w.o. cc & (c) Ours \\ 
    \end{tabular}
    \vspace{-0.2cm}
    \caption{Clustering of visual samples within the color superordinate. We compare NSCL~\cite{mao2019neuro} with our model, and ours without concept clustering. The color values used in (a)$\sim$(c) are exactly those used in the CLEVR-Perturb test set (please refer to Figure~\ref{fig:color} for more details).}
    \label{tab:cluster}
\end{figure*}

\subsection{Clustering within Superordinate}
For the CLEVR-Perturb test set, we compare in detail the clusters formed by the visual samples around different concepts. Clustering results are shown in Figure~\ref{tab:cluster}. We observe that even the variant without concept clustering learns a clearly more discriminant subspace than NSCL~\cite{mao2019neuro}. Equipped with concept clustering, our model further enhances the discrimination in the superordinate. The better clustering results also account for higher reasoning accuracy. Note that, for fairness, in the comparison in Table~\ref{tab:perturb}, we avoid querying about \textit{color} on the CLEVR-Perturb test set.
Furthermore, Figure~\ref{tab:cluster} (a)$\sim$(c) demonstrate that our model improves the discrimination of different primitive concepts (\eg `gray', `cyan', and `yellow') within the `color' superordinate, even when the input colors are perturbed.

\begin{figure}[t]
\centering
\includegraphics[width=.6\textwidth]{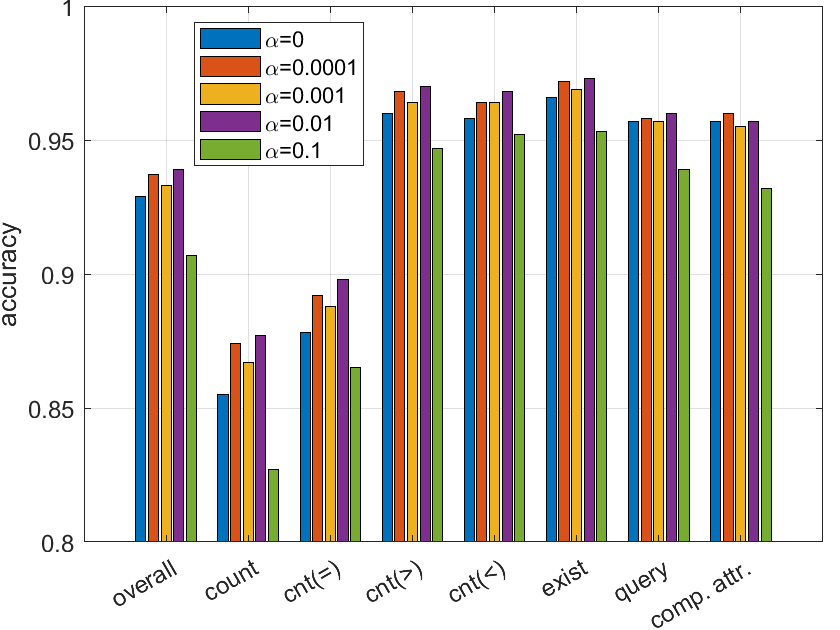}
\caption{Question answering accuracy \textit{w.r.t} the decay coefficient $\alpha$ that ranges from $0.0001$ to $0.1$ in logarithmic scale. Performance of the proposed model without concept clustering serves as a reference.}
\label{fig:alpha}
\end{figure}

\subsection{Sensitivity Analysis}

Figure~\ref{fig:alpha} presents the sensitivity of the proposed model \textit{w.r.t} the decay coefficient $\alpha$ in Eq.~(\ref{eq:contra}). The cached samples around a concept provide more visual information for reference in identification. We tune $\alpha$ from $0.0001$ to $0.1$ in a logarithmic scale, and record the accuracy change in each question type. The results show that all the variants perform better on binary reasoning questions, including comparing numbers or attributes, reasoning existence, and querying attributes, than answering specific counting questions. We analyze that for the learner, counting is a more complicated discretization process where probabilities higher than a threshold are binarized and then added up. Not surprisingly, this complicated task benefits the most from improving the learner's robustness, both by the proposed superordinate abstraction and by the concept clustering. Figure~\ref{fig:alpha} illustrates that as the increase of $\alpha$, the performance is improved first and then becomes worse than the baseline. It indicates that there is a trade-off between the effect of linguistic tokens and visual samples within a superordinate.

\subsection{Reasoning Details}
\begin{figure}[h]
    \centering
    \includegraphics[width=0.96\textwidth]{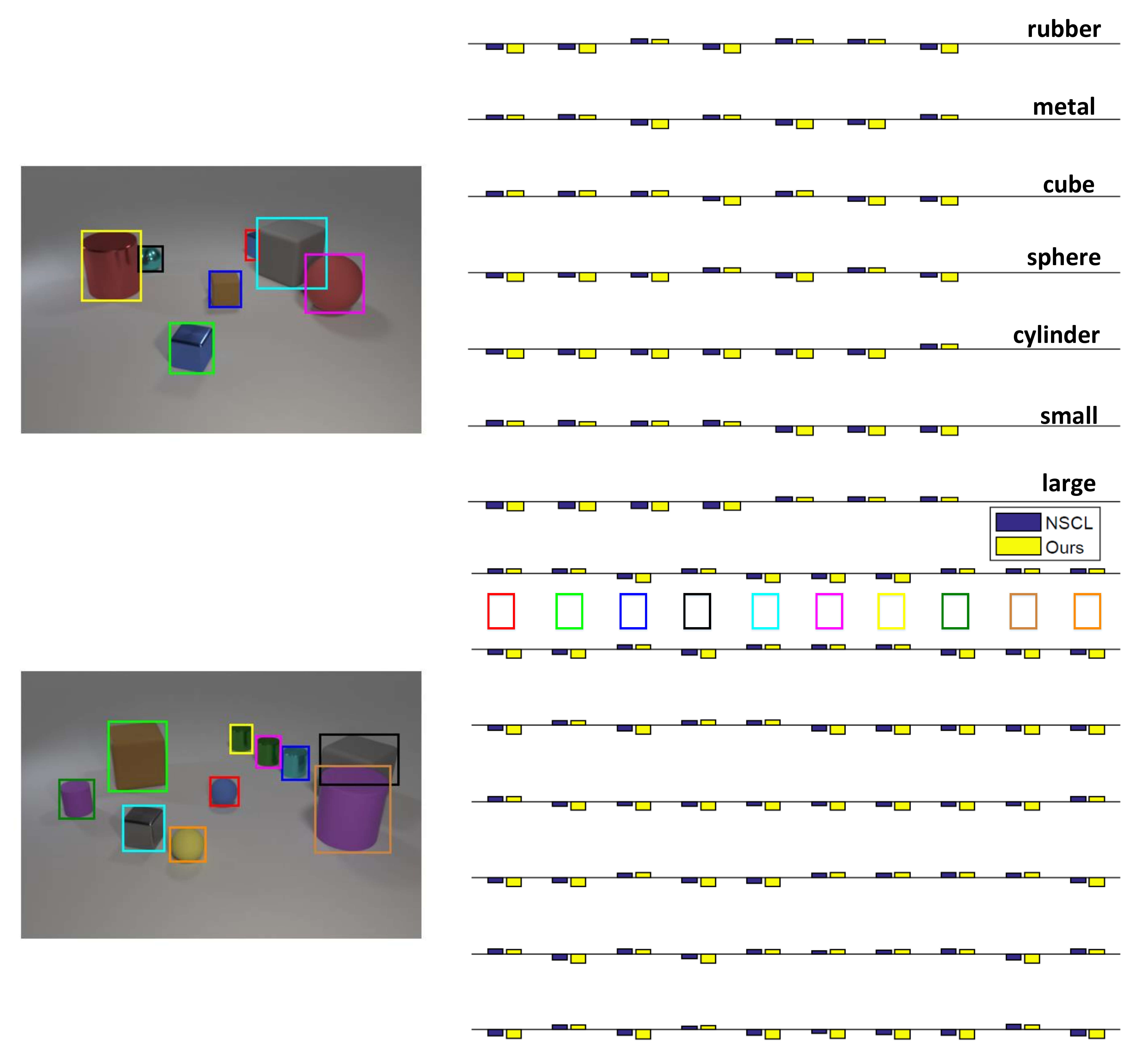}
    \caption{Illustration of reasoning details on the CLEVR-CoGenT val-A set. The colorful rectangles refer to the object in corresponding bounding boxes.}
    \label{fig:perturb_supp}
\end{figure}

\begin{figure}
    \centering
    \includegraphics[width=0.96\textwidth]{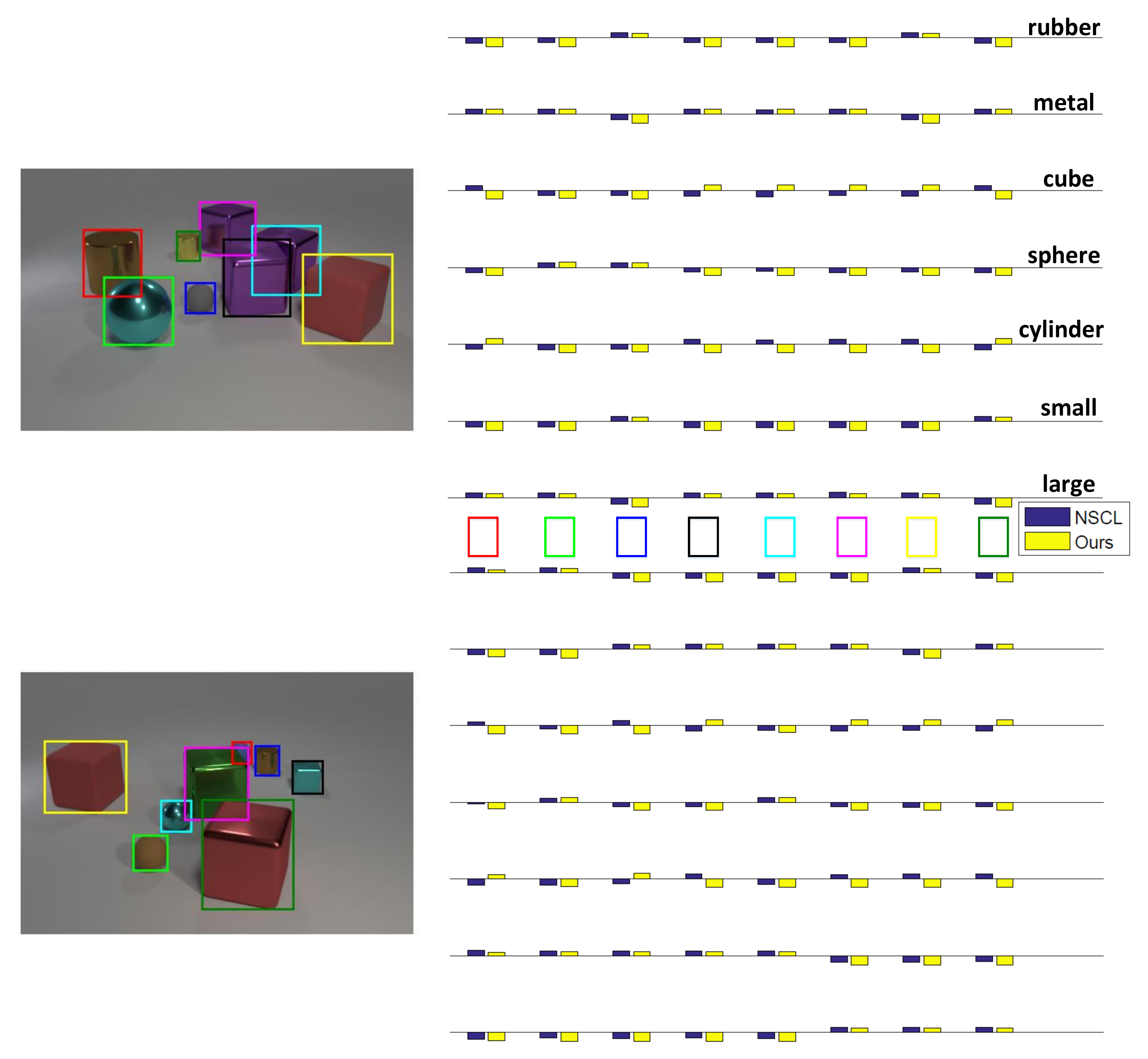}
    \caption{Illustration of reasoning details on the CLEVR-CoGenT val-B set. The colorful rectangles refer to the object in corresponding bounding boxes.}
    \label{fig:cogent_supp}
\end{figure}

To provide an intuitive comparison between our model and NSCL~\cite{mao2019neuro}, we list more details of the reasoning details on the val-A and the val-B splits of CLEVR-CoGenT dataset~\cite{johnson2017clevr} in Figure~\ref{fig:perturb_supp} and Figure~\ref{fig:cogent_supp}, respectively. It shows that NSCL~\cite{mao2019neuro} perform almost perfectly on val-A, but mixes `cube' and `cylinder' on val-B. Our model overcomes the bias on val-B split.

\section{Conclusion}
In this paper, we propose a visual superordinate abstraction framework for concepts learning. The learner acquires linguistic abstraction from soft-aligned questions and contributes to the discrimination of visual abstraction. On top of the framework, we devise a quasi-center concept clustering and a superordinate shortcut learning schemes to address such issues as perturbations and biased training. Experiments under different settings verify the superiority of the proposed model. 

A potential limitation of this paper is the proposed visual superordinate abstraction has not been validated on large-scale real-world datasets. The synthesized CLEVR datasets provide ideal and controllable environments that allow the community to directly explore and evaluate different concept learners. In this paper, we mainly focused on analyzing the potential bottleneck of existing methods, and pinpointed that most of them ignored the valuable abstraction capability in human reasoning. In future works, we aim to further explore the capacity of the proposed visual superordinate abstraction in real-world scenarios. 


\bibliographystyle{plain}
\bibliography{references}

\end{document}